\documentclass[letterpaper, 10 pt, conference]{ieeeconf}  

\IEEEoverridecommandlockouts                              

\usepackage{graphics} 
\usepackage{epsfig} 
\usepackage{mathptmx} 
\usepackage{times} 
\usepackage{amsmath} 
\usepackage{amssymb}  
\usepackage[nolist,nohyperlinks]{acronym}
\usepackage{hyperref}

\usepackage{bm}
\usepackage{caption} 
\usepackage{subcaption}
\usepackage{graphicx}
\usepackage{makecell}
\usepackage{algorithm}
\usepackage{algpseudocode}
\usepackage{wrapfig}
\usepackage[nolist,nohyperlinks]{acronym}
\usepackage{units}
\usepackage{mathtools}
\usepackage{multirow}
\usepackage{microtype}
\usepackage{xcolor}

\title{\LARGE \bf
Unified Data Collection for Visual-Inertial Calibration via Deep Reinforcement Learning
}

\author{Yunke Ao$^{*}$, Le Chen$^{*}$, Florian Tschopp, Michel Breyer, Andrei Cramariuc, Roland Siegwart  

\thanks{*Authors contributed equally and both can be considered as
first authors.}

\thanks{$\, $ Authors are with the Autonomous Systems Lab, ETH Zurich, Leonhardstrasse 21, 8092 Zurich, Switzerland. Email:
    {\tt\footnotesize yunkao,lechen@ethz.ch}}%
        
}

\begin{acronym}
\acro{mpc}[MPC]{model predictive control}
\acro{rl}[RL]{reinforcement learning}
\acro{pso}[PSO]{particle swarm optimization}
\acro{mdp}[MDP]{Markov decision process}
\acro{pomdp}[POMDP]{partially observed Markov decision process}
\acro{vi}[VI]{visual-inertial}
\acro{imu}[IMU]{inertial measurement unit}
\acro{fov}[FoV]{field of view}
\acro{sgd}[SGD]{stochastic gradient decent}
\acro{sac}[SAC]{Soft Actor-Critic}

\end{acronym}

\newcommand{\etal}{\textit{et al}. }

\begin{document}

\maketitle

\thispagestyle{empty}
\pagestyle{empty}

\begin{abstract}
Visual-inertial sensors have a wide range of applications in robotics. However, good performance often requires different sophisticated motion routines to accurately calibrate camera intrinsics and inter-sensor extrinsics. This work presents a novel formulation to learn a motion policy to be executed on a robot arm for automatic data collection for calibrating intrinsics and extrinsics jointly. Our approach models the calibration process compactly using model-free deep reinforcement learning to derive a policy that guides the motions of a robotic arm holding the sensor to efficiently collect measurements that can be used for both camera intrinsic calibration and camera-IMU extrinsic calibration. Given the current pose and collected measurements, the learned policy generates the subsequent transformation that optimizes sensor calibration accuracy. The evaluations in simulation and on a real robotic system show that our learned policy generates favorable motion trajectories and collects enough measurements efficiently that yield the desired intrinsics and extrinsics with short path lengths. In simulation we are able to perform calibrations $10\times$ faster than hand-crafted policies, which transfers to a real-world speed up of $3\times$ over a human expert. 
\footnote[1]{The code of this work is publicly available at:~\url{https://github.com/ethz-asl/Learn-to-Calibrate/tree/master}.}

\end{abstract}

\section{INTRODUCTION}

In recent years, \ac{vi} sensors, which include one or more cameras and an \ac{imu}, have become a key component of many autonomous mobile robot systems since they can provide precise motion estimates and are well suited for use in various high-level robotic tasks~\cite{bloesch2017iterated,8421746,8633393,leutenegger2015keyframe,corke2007introduction}.
%
%
Precise calibration, which for \ac{vi} sensors refers to the parameters for the camera intrinsics, the camera-IMU extrinsics, and the time offset between the different sensors, are of great importance to the accuracy and performance of \ac{vi} systems~\cite{nikolic2014synchronized,tschopp2020versavis,kelly2011visual}.
%
%
However, it is usually non-trivial to perform the calibration by hand or with a manually pre-programmed operator, since it often requires complex motion routines ensuring observability in controlled environments~\cite{kelly2011visual,nobre2019learning,furgale2013unified,furgale2012continuous}.
In order to guarantee high calibration accuracy, typically large amounts of data are recorded for optimization. 
But the calibration time for the trajectory quickly grows as more measurements are collected, which makes calibration a quite time-consuming process~\cite{scaramuzza2019visual}. 
In addition, given the current pose and calibration data collection state, the movement that renders the best calibration results effectively and efficiently for the next step is not obvious. 
%

In early studies, several works applied \ac{rl} to get the best sequence of trajectories for \ac{vi} calibration.
Nobre \etal\cite{nobre2019learning} proposed using Q-learning to select a sequence of motions from a predefined library aiming to render sufficient observability of the calibration problem.
However, this approach only suggests which predefined motions to choose from the empirically designed library without exploring new possible motion primitives. 
To look into new possible trajectories, Chen~\etal\cite{chen2020learning} modeled the calibration process as a \ac{mdp} and proposed a model-based deep \ac{rl} calibration framework with an adapted version of \ac{pso}~\cite{kennedy1995particle} for sampling to generate different trajectories for intrinsic and extrinsic calibration.
%
However, the \ac{mdp} in this method suffers from high dimensionality and the learned open-loop trajectories do not adapt to the feedback of the current state.

To deal with the shortcomings of current systems (see Section~\ref{sec:rw}), we model calibration as an \ac{mdp} in a compact way and use model-free deep \ac{rl}~\cite{haarnoja2018soft} to learn a policy that guides the movements of the \ac{vi} sensors. A robot arm then executes the motions. According to the current pose and collected measurements, the policy generates the next transformation that optimizes calibration accuracy and efficiency.
Additionally, we simulate different \ac{vi} sensor configurations including symmetric and asymmetric distortions to model the variability that also exists in the real world, enabling better transferability to real scenarios.
Compared with other solutions, the action space of our model has fewer constraints, making it possible to learn a more general policy for calibration.
Moreover, our method obtains only one policy that can be used both for intrinsic and extrinsic calibration.
The main contributions of this work are as follows:
\begin{itemize}
\setlength\itemsep{0em}

\item To the best of our knowledge, we are the first to use model-free deep \ac{rl} to learn a policy that generates trajectories for efficiently collecting calibration measurements.


\item We propose a novel formulation to jointly calibrate both camera intrinsic and \ac{vi} extrinsic parameters with only one trajectory.

\item The simulated evaluation shows that the trajectories generated by the learned policy deliver more accurate and efficient calibrations compared with other trajectories. The real experiments confirm the transferability of our approach to real scenarios.
\end{itemize}

\section{RELATED WORKS}

\label{sec:rw}

\begin{figure*}[t]
    \centering
      \vspace{0.1cm}
      \setlength{\belowcaptionskip}{-0.5cm} 
    \includegraphics[width=0.8\linewidth]{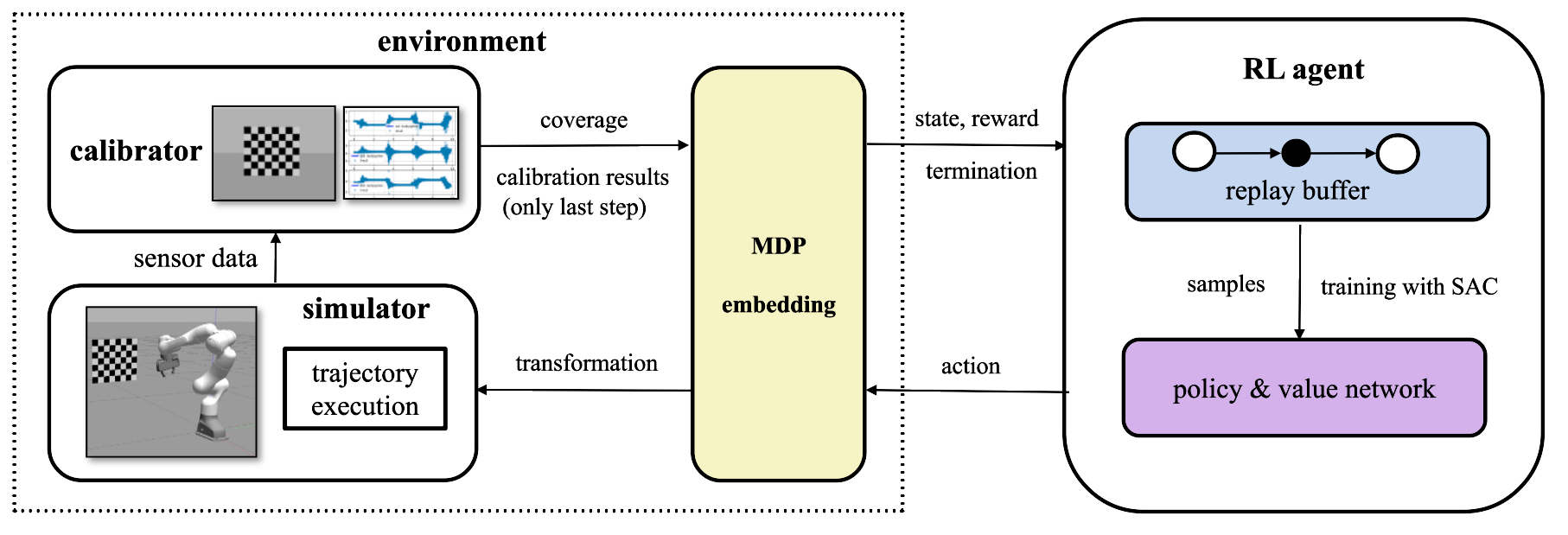}
    \caption{Entire framework of our method. %
At each step, the agent first proposes an action based on the current state and policy.
This action is then used to compute the transformation of the pose of the end-effector.
Given the next pose, the simulation platform executes the trajectory and records the resulting sensor data.
The calibrator then updates the data acquisition status based on the sensor measurements.
If the step is the last step, the calibrator will also perform a final calibration.
All the results and data acquisition status are then transformed into new states and rewards by the \ac{mdp} embedding module, which is recorded in the replay buffer of the \ac{rl} agent.
The calibration, simulation, and \ac{mdp} embedding modules can be jointly regarded as the environment for the \ac{rl} learner to interact.
Apart from generating new interactions at each iteration, the agent also samples recorded interaction data from the replay buffer to update the policy using the \ac{sac} algorithm~\cite{haarnoja2018soft}. }
    \label{fig:framework}
\end{figure*}

Calibrating a \ac{vi} system can be divided into two parts: camera calibration and camera-IMU calibration.
The goal of camera calibration is to determine the camera intrinsics and the distortion model coefficients~\cite{andrew2001multiple}.
Camera-IMU calibration implies obtaining the camera-IMU extrinsics and the time offset between different sensors~\cite{nikolic2014synchronized,tschopp2020versavis}. 
The classical method for camera calibration is to construct constraints using a known calibration pattern and formulate an optimization problem to determine the parameters~\cite{sturm1999plane,zhang2004extrinsic}.
For camera-IMU extrinsic calibration, the most popular and reliable methods also require a calibration board.
Furgale~\etal\cite{furgale2013unified,furgale2012continuous} proposed to parameterize the pose and IMU bias trajectories using B-splines and applied continuous-time batch estimation to jointly calibrate the time offsets and spatial transformations between multiple sensors.
Those methods are able to retrieve the calibration parameters of the sensors with good accuracy. However, the optimal trajectories for measurement collection in terms of efficiency, observability, and accuracy, remain unknown. Calibration still requires expert knowledge to collect enough effective data and is often a time-consuming process for inexperienced people.

In recent years, automatic calibration of \ac{vi} sensors has become an active research topic in robotics communities.
The approaches for automatic calibration can be divided into two categories: trajectory optimization based and \ac{rl} based.
Most of the trajectory optimization based methods focus on optimizing the observability Gramian of the trajectories on the calibration parameters~\cite{hausman2017observability,preiss2017trajectory}.
However, these methods ignore some practical requirements of calibration, such as minimizing path length, because they are difficult to model.

\ac{rl} based methods can better include those general requirements.
%
%
Nobre \etal\cite{nobre2019learning} defined a library containing different motions and applied Q-learning to select a sequence of motions from the library with the goal to obtain enough observability for calibration.
%
%
However, the performance, as well as efficiency of calibration, is constrained by the predefined library because many possible motion sequences are not included or explored.

%
Chen~\etal\cite{chen2020learning} addressed this problem by learning continuous parameters of calibration trajectories using a model-based \ac{rl} algorithm with adapted \ac{pso}.
%
%
This work trained two networks, one for intrinsic and one for extrinsic calibration.
They define the action at each time as a looped trajectory which is described by 36 parameters.
%
%
The state is defined as the concatenation of all actions and calibration results of previous time steps.
Its number of dimensions thus grows with increasing time steps, resulting in a high dimensional \ac{mdp} that is difficult to solve.
In addition, the proposed method learns open-loop trajectories which do not adapt to the feedback of the current state.
%

In contrast, our proposed approach models the calibration process in a compact way. We define an action as the transformation to the next pose, which only requires 6 dimensions. And the state, consisting of the information status and the current pose, now has a fixed number of dimensions. Thanks to the compactness and efficiency of the proposed \ac{mdp}, we are able to apply a model-free algorithm~\cite{haarnoja2018soft} to learn policies.
%
%
Moreover, our current approach includes symmetric as well as asymmetric distortion in the experiment and trains only one network which can be used for both intrinsic and extrinsic calibration.


\section{METHOD}

\label{sec:method}

\begin{figure*}[t]
    \centering
      \setlength{\belowcaptionskip}{-0.5cm} 
    \includegraphics[width=0.8\linewidth]{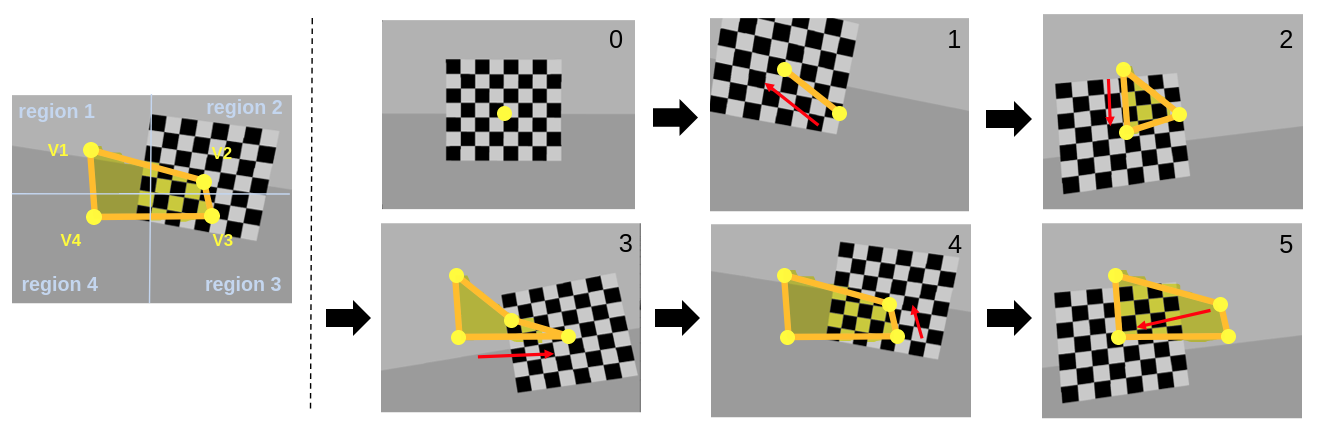}
    \caption{Illustration of image coverage. The left image shows the division of the 4 regions of the image view and their corresponding vertices. The right part shows an example of propagation of image coverage with increasing steps. The orange polygon defined by the four yellow vertices always covers the positions swept by the target board. The red arrows show the approximated movement of the target board in the image view during each step.}
    \label{fig:image_coverage}
\end{figure*}%

Our goal is to generate a sequence of direct transformations that form a trajectory for the end-effector with the \ac{vi}-sensor fixed on it, based on which accurate intrinsic and extrinsic parameters can be calibrated using the acquired sensor data.
We solve this problem by learning a policy for generating the step-by-step transformation based on feedback information at each step.
The whole proposed learning framework is shown in Figure \ref{fig:framework}, including the simulator, calibrator, \ac{mdp} embedding module, and the \ac{rl} agent. 

\subsection{Visual-Inertial Calibration}
\label{sec:vi_calibration}
In our method, both camera intrinsic and camera-IMU extrinsic calibration are performed using the \textit{Kalibr}~\cite{furgale2013unified, furgale2012continuous} framework. 
In \textit{Kalibr}, calibration is formalized as a large-scale optimization problem, using the Levenberg-Marquardt algorithm, to minimize the error between the predicted and obtained measurements.
%
%
Other outputs from the framework also provide useful information for estimating the calibration quality, e.g. reprojection error distribution, motion bias curve, etc.
%
%
In each episode of interaction for the agent, we only call \textit{Kalibr} once at the last step if the data acquisition status fulfills our predefined requirements, in order to reduce the computational cost for training.

\subsection{MDP Model for Calibration}

\label{sec:mdp}

\paragraph{Action Definition}
As the whole trajectory consists of a sequence of direct transformations, we define the action of each step as a translation and rotation of the pose of the end-effector $A = [\rho, \theta, \phi, \alpha, \beta, \gamma]$. 
Here $\rho$ is the length of translation, $\theta$, and $\phi$ are azimuth and polar angles indicating the direction of translation. 
$\alpha$, $\beta$, and $\gamma$ are Euler angles that determine the rotation.
We also set constraints on $\rho$, $\alpha$, $\beta$, and $\gamma$ to guarantee feasible motions.
%

\paragraph{State Definition}
Regarding the problem formulation of calibration, the final calibration results depend on all the data acquired from the whole trajectory.
%
%
In~\cite{chen2020learning}, all the trajectories and calibration outputs of the previous steps are used as a description of the status of the data acquired so far.
%
To simplify the modeling and reduce the difficulty of learning, we only use a fixed number of variables that define the range of the previously acquired data as our state, rather than using the whole history.

For image data, we use the coverage of the center position, change in size, and skew of the target board in the image view to indicate the calibration progress.
For the coverage of the target board center position, instead of only recording the maximum and minimum value of X and Y coordinates of the center position, we use a polygon with four vertices to describe the 2D range in a more specific way.
As is shown in Figure \ref{fig:image_coverage}, we divide the image equally into four parts: left-up, right-up, right-down and left-down, denoted by the $\mathcal{R}_1,\mathcal{R}_2, \mathcal{R}_3,$ and $\mathcal{R}_4$ respectively.
In each region $\mathcal{R}_i$, we place one vertex $V_i$ that has to stay inside the respective region.
Given all the positions that the center of the target board has covered $\{[u_j, v_j]\}_{j=1}^N$, the update rules for the vertices at each step are:

\begin{align}
    V_1 = \begin{bmatrix}
    \min\limits_{[u_j, v_j]\in \mathcal{R}_1} u_j, 
    \min\limits_{[u_j, v_j]\in \mathcal{R}_1} v_j
    \end{bmatrix}, &
    V_2 = \begin{bmatrix}
    \max\limits_{[u_j, v_j]\in \mathcal{R}_2} u_j,
    \min\limits_{[u_j, v_j]\in \mathcal{R}_2} v_j
    \end{bmatrix} \ \notag\\
    V_3 = \begin{bmatrix}
    \max\limits_{[u_j, v_j]\in \mathcal{R}_3} u_j,
    \max\limits_{[u_j, v_j]\in \mathcal{R}_3} v_j
    \end{bmatrix}, &
    V_4 = \begin{bmatrix}
    \min\limits_{[u_j, v_j]\in \mathcal{R}_4} u_j,
    \max\limits_{[u_j, v_j]\in \mathcal{R}_4} v_j
    \end{bmatrix} \
\end{align}

%
%
%
The range of the target board sizes is encoded by the minimum and maximum proportion of the area of the target board in the image view, denoted by $A_{min}$ and $A_{max}$. 
For skew coverage, we use the range of angles between two fixed edges of the target board in the image view $\eta_{min}$ and $\eta_{max}$.

To encode the range of motion, we use the maximum differences between the current and the next poses in the previous steps $\Delta p_{max}, \Delta \Theta_{max}$, where $p$ and $\Theta$ denote the position and orientation of the sensor, respectively.
This corresponds to the span of acquired IMU sensor data because larger differences will results in larger velocities and accelerations for direct transformations with the default controller. 
For orientation, we approximate the difference as the difference of Euler angles.
Note that the coverages are all normalized between 0 and 1 to benefit training and terminal condition design.
In addition to the range of acquired data, we also include the current pose of the end-effector $p, \Theta$ in the state.
In this way, the full state of data acquisition is defined as: 
\begin{equation}
\begin{aligned}
    S_t = [V_1^t, V_2^t, V_3^t, V_4^t, A_{min}^t, A_{max}^t, \eta_{min}^t, \eta_{max}^t,\qquad \quad \ \\ \Delta p_{max}^t, \Delta \Theta_{max}^t, p^t, \Theta^t] \in \mathbb{R}^{24}.
\end{aligned}
\end{equation}

%
\paragraph{Terminal Condition}
In contrast to the approach proposed by Chen \etal~\cite{chen2020learning}, we additionally define a terminal condition to assist the agent in learning a desired policy that could generate trajectories with fewer steps.
This could also alleviate unnecessary further data collection and reduce the time cost for calibration

%
The MDP process is terminated if all the coverage parameters have reached predefined thresholds, and an additional terminal reward is given.
This design is intended to encourage the agent to maximize the coverage of all properties, rather than only extremely cover one or two of them.
Although the values for the thresholds are set to be not too difficult to achieve to stabilize the learning process, in practice the agent will also learn to achieve as much coverage as possible by maximizing the total rewards.
%

\paragraph{Reward Design}
The rewards include step-wise rewards $R_t$ and a terminal reward $R_T$. 
For the step-wise rewards $R_t$, we assign a positive reward for increased coverage and a negative reward for higher path length.
Here we include path length because the shortness of the calibration sequence heavily reduces the run time of the calibration software that is a non-linear optimization process.
%
%
For the step-wise reward we have a weighted sum
\begin{equation}
\begin{aligned}
    R_t = \|S_t[0:18] - S_{t-1}[0:18]\|_1 \\- c_1\|p^t-p^{t-1}\|_2 & - c_2 \|\Theta^t-\Theta^{t-1}\|_2,
    \label{eq:step_r}
\end{aligned}
\end{equation}
where $c_1$ and $c_2$ are tunable positive hyperparameters. The terminal reward $R_T$ is designed to favor behavior that reaches the terminal condition. In our case, $R_T$ includes a positive constant reward and dynamic rewards for accurate and less uncertain calibration results.
Given the ground truth and estimated calibration parameters $\Phi$ and $\hat{\Phi}$ respectively, the terminal reward is defined as
\begin{align}
    R_T = c_3 + \frac{c_4\|\Phi\|_2 }{\|\Phi-\hat{\Phi}\|_2},
    \label{eq:terminal_r}
\end{align}
where $c_3$ and $c_4$ are tunable positive hyperparameters. 
High terminal rewards are given for low calibration errors to encourage the agent to learn accurate calibrations.

\subsection{Soft-Actor Critic Reinforcement Learning}

In our framework, we apply \ac{sac}~\cite{haarnoja2018soft} to train the agent to learn the optimal policy.
During the online training process, our agent jointly learns a policy network, a Q-network, and a value network together.
The parameters of networks are updated using the maximum entropy framework, which introduces entropy into the original actor-critic update to improve exploration of the environment.
%
Furthermore, this framework also enables off-policy learning that improves the sample efficiency.
Due to the high computational cost for calibration, we chose \ac{sac} as it requires few experiences to reach high performance.

\section{EXPERIMENTS}
\label{sec:experiment}

\subsection{Evaluation with real platform}
\begin{figure*}[t]
    \centering
  \setlength{\belowcaptionskip}{-0.5cm} 
    \includegraphics[width=0.95\linewidth]{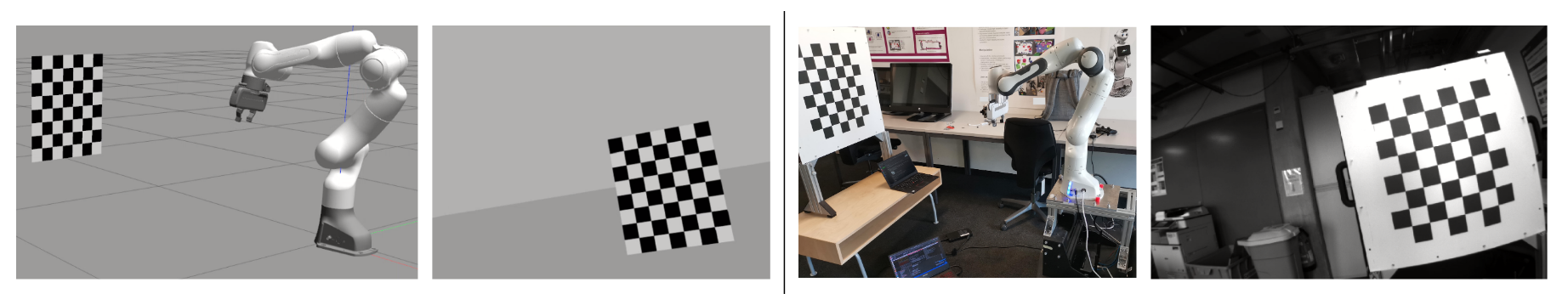}
    \caption{Simulation (left) and real experiment setup (right) for \ac{vi} calibration. The example images from the camera sensor are shown respectively.}
    \label{fig:real}
\end{figure*}

In this section, we conduct experiments to train and evaluate our learned policy. 
The whole pipeline includes policy training, simulation evaluation, and validation on a real platform.
The results show that our learned policy performs best among all the candidate trajectories and policies for both intrinsic and extrinsic calibration.
We also verify that our policy can be applied on a real robot arm to calibrate a real \ac{vi}-sensor.
\subsection{Policy Training}
\label{sec:policy}
We use Gazebo~\cite{koenig2004design} for dynamics and sensor simulation. 
Our simulation environment includes a checkerboard and a VI sensor consisting of a pinhole camera and an IMU mounted on the end-effector of a FRANKA EMIKA Panda robot arm\footnote{https://erdalpekel.de/?p=55}.
%
%
The detailed sensor settings are shown in Table \ref{tab:IMU} and \ref{tab:camera}. 
We include noise and drift for the IMU and distortion for the camera to achieve a more realistic simulation.
During training, the parameters for camera intrinsics and camera-IMU extrinsics are re-sampled from Gaussian distributions after each episode, to ensure that the model learns to generalize well also to other similar sensors.
%
%
Our target board is a $7\times6$ checkerboard with $\unit[7]{cm}\times\unit[7]{cm}$ squares.
The distance from the target to the robot arm's initial pose is $\unit[1.5]{m}$.
In each step, the direct transformation is executed by the robot arm using the \textit{MoveIt} motion planning package~\cite{chitta2012moveit}.
If the terminal condition is satisfied at the last step, both intrinsic and extrinsic calibration will be performed using \textit{Kalibr}, where the intrinsic result is used as input for the extrinsic calibration.


\begin{table}[!t]
\footnotesize
    \centering
    \vspace{0.2cm} 
    \begin{tabular}{c | c}
    \hline
    \textbf{Parameters} & \textbf{Values} \\
    \hline
        frequency & 200 Hz \\
        accelerometer drift & $\unit[0.006]{m/s^2}$ \\
        accelerometer noise & $\unit[0.004]{m/s^2}$\\
        gyroscope drift & $\unit[0.000038785]{rad/s}$\\
        gyroscope noise & $\unit[0.0003394]{rad/s}$\\
         \hline
    \end{tabular}
    \caption{Simulation settings for the IMU.}
    \label{tab:IMU}
\end{table}

\begin{table}[!t]
\footnotesize
    \centering
    \vspace{-0.2cm} 
    \begin{tabular}{c | c}
    \hline
    \textbf{Parameters} & \textbf{Values} \\
    \hline
        frequency & 10 Hz \\
        width & $\unit[640]{px}$ \\
        height & $\unit[480]{px}$\\
        nominal horizontal FOV & $\unit[1.0]{rad}$\\
         model & pinhole-radtan\\
         \hline
    \end{tabular}
    \caption{Simulation settings for the camera.}
    \label{tab:camera}
\end{table}


\begin{table}[!t]
\footnotesize
\centering
\vspace{-0.2cm} 
\setlength{\belowcaptionskip}{-0.4cm}
\scalebox{0.95}{
\begin{tabular}{c|c|c c}
\hline
& \textbf{Parameters} & \textbf{Mean} & \textbf{Std.} \\
\hline
\textbf{Intrinsics}  & FOV [rad] & 1.00 & 0.05\\ 
\hline
\multirow{ 2}{*}{\textbf{Distortion}}   & $k_1$, $k_2$ & 0.00  & 0.02 \\
& center & 0.00  & 0.05 \\
\hline
\multirow{ 6}{*}{\textbf{Extrinsics}} & X {[}m{]} & 0.06 & 0.01 \\    
& Y {[}m{]} & 0.00 &0.01 \\
&Z {[}m{]} &-0.10  &0.01  \\
& Roll {[}rad{]}&0.00 & 0.10  \\
&Pitch {[}rad{]} &0.00 & 0.10\\
&Yaw {[}rad{]} & 1.57& 0.10\\
\hline
\end{tabular}}
\caption{Gaussian distribution settings for intrinsics and extrinsics parameters during training. The variances are within tolerance to specialize to one sensor model (e.g., for high throughput factory calibration), and in Section~\ref{sec:experiments-real} we show that our model also works on out of distribution parameters.}
\label{tab:dist}
\end{table}


%
With these environment settings and predefined \ac{mdp}, we perform \ac{sac} to learn the optimal policy for calibration.
For the value network, Q-network, and policy networks of the agent, we use fully connected networks with 2 hidden layers and a size of 256.
We restrict the maximal number of steps for each episode to be 20 and train the agent for nearly 15000 steps. 
At that stage, the performance of the agent converges, and it can achieve the terminal condition for most of the episodes.

\subsection{Evaluation in Simulation}
In the simulation evaluation experiments, we test 6 different types of policies or trajectories for 3 calibration tasks: pure intrinsic calibration, pure extrinsic calibration with known intrinsic ground truth, and calibration of both intrinsic and extrinsic parameters.
For comparison, we obtained baselines that include handcrafted trajectories for intrinsic and extrinsic calibration from expert knowledge (a short and a long version) and trajectories learned with model-based \ac{rl}~\cite{chen2020learning}.
In addition, our learned policy is also compared with randomly parametrized trajectories introduced in~\cite{chen2020learning} and a random policy of direct transformations that parametrizes movement similarly to our proposed approach.

\paragraph{Intrinsic calibration}
%
The intrinsics are calibrated using the \textit{Kalibr}~\cite{furgale2013unified} toolbox, with a pinhole camera projection model and a radial-tangential distortion model.
For each policy and trajectory we randomly sample 20 camera settings under the same Gaussian distribution specified in Section~\ref{sec:policy}, except for the center of distortion.
For testing we keep the center of distortion fixed as $[0.5, 0.5]$ to ensure the ground truth of the camera parameters is known.

The results are shown in Table \ref{tab:intrinsic}, in which we compare the average path length, calibration time and intrinsic calibration error.
The results show that our policy, learned with model-free \ac{rl}, achieves close performance with the long hand-crafted trajectory, that is more than 4 times longer and requires 3 times more time to calibrate.
The short hand-crafted trajectory that has a more comparable length to our method performs significantly worse, while still taking double the time to calibrate.

We can also see that the random direct moving policy, in which we randomly pick a pose within the constraints to move to at each time step, achieves a high performance compared with other trajectories.
This could be because the random moving policy is less constrained than the previous baselines which had to perform cyclical motions and could therefore not easily achieve such a good coverage.
In addition, the baselines proposed by Chen~\etal~\cite{chen2020learning} were trained without randomizing the distortion model, possibly causing them to have lower generalizability.

%
\paragraph{Extrinsic calibration with known intrinsic ground-truth}
We first test the extrinsic calibration performance separately by using known camera intrinsic ground truth.
%
We mainly compare the extrinsic calibration accuracy, the variance of calibration error, path length, and calibration time.
Same as before, we evaluate each policy with 20 randomly sampled intrinsics and extrinsics with the same distribution used for training except for the distortion center, which we keep fixed.
After each resampling of sensor settings, the true sensor intrinsic parameters are input directly to \textit{Kalibr}.

The results are shown in Table \ref{tab:extrinsic}. 
Our learned policy using model-free \ac{rl} outperforms all the baselines on extrinsic calibration accuracy.
Our path length is not the shortest in this case, but our policy is jointly trained also for intrinsic calibration which requires higher coverage.
Even so, the calibration time for our proposed policy is the best, most likely due to the simpler motion patterns of a direct movement policy which makes the optimization problem easier.
Model-based learned trajectory and random moving policy have similar performance in this task, and they are both better than predefined trajectories.
The poor performance of hand-crafted policies is simply a matter of time, given enough time to collect data any sane calibration policy will eventually reach the same optimal result.

%
\paragraph{Joint intrinsic and extrinsic calibration}
Finally, we evaluate the joint intrinsic and extrinsic calibration performance for each policy.
Specifically, for each category of policies to test, we use the policy trained for intrinsic and extrinsic calibration to calibrate the intrinsics and extrinsics respectively, where the calibrated intrinsic is input to the extrinsic calibration step.
As for our learned policy, we just use one single policy for both calibrations.
Here, we include the randomization of the center of distortion as well as asymmetric distortion as we do not depend on ground truth intrinsics.
We evaluate the performance of policies only based on their extrinsic calibration results.
This is because the ground truth for intrinsic is not known if asymmetric distortions are introduced in the simulation.
Finally, the accuracy of intrinsic results also influences extrinsic calibration. Therefore, the final extrinsic calibration accuracy corresponds to the overall evaluation of both intrinsic and extrinsic calibration.

The results of the simulation evaluation are shown in Table~\ref{tab:extrinsic_intrinsic}.
%
%
Our policy obtains the lowest extrinsic calibration error with the lowest total path length and calibration time, even though we use the same trajectory for both intrinsic and extrinsic calibration.
%
%
In addition, it is interesting to see that when considering asymmetric distortions, the hand-crafted trajectory also achieve nearly the highest accuracy.
%
What we show is that different calibration policies can reach an optimal result given enough time, but our proposed policy is able to learn better motion primitives that human experts.
We are able to do in less than 5 minutes what would take a human more than 50 minutes.


\begin{table*}[!htp]
\footnotesize
\centering
\vspace{0.15cm} 
\setlength{\belowcaptionskip}{-0.2cm}
\begin{tabular}{c|cccc}
\hline
\textbf{Intrinsic calibration}   & \textbf{Mean error {[}\%{]}} & \multicolumn{1}{l}{\textbf{Error std {[}\%{]}}} & \textbf{Path length {[}m{]}} & \textbf{Calibration time {[}s{]}}\\ \hline
random trajectory                & 14.81            & 7.27             & 2.27             & 67 \\
hand-crafted trajectory (long)   & \textbf{3.27}    & \textbf{1.90}    & 4.05             & 148  \\
hand-crafted trajectory (short)  & 8.79             & 2.39             & 2.77             & 63 \\
model-based learned trajectory   & 7.65             & 7.59             & 2.04             & 133  \\
random moving policy             & 6.60             & 6.33             & 1.16             & 44 \\
\textbf{model-free learned policy (ours)} & 3.90             & 2.83             & \textbf{0.86}    & \textbf{39} \\ \hline
\end{tabular}
\caption{Evaluation results of different trajectories and policies for intrinsic calibration.}
\label{tab:intrinsic}
\end{table*}

\begin{table*}[!htp]
\footnotesize
\centering
\setlength{\belowcaptionskip}{-0.2cm}
\begin{tabular}{c|cccc}
\hline
\textbf{Extrinsic calibration} & \textbf{Mean error {[}\%{]}} & \multicolumn{1}{l}{\textbf{Error std {[}\%{]}}} & \textbf{Path length {[}m{]}} & \textbf{Calibration time {[}s{]}}\\ \hline
random trajectory                 & 9.48           & 6.87            & 2.27           & $1908$ \\
hand-crafted trajectory (long)    & 1.80           & 1.06            & 4.05           & $3019$ \\
hand-crafted trajectory (short)   & 2.66           & 2.14            & \textbf{0.56}  & $611$            \\
model-based learned trajectory    & 1.76           & 1.19            & 1.20           & $756$            \\
random moving policy              & 1.73           & 1.34            & 1.16           & $569$            \\
\textbf{model-free learned policy (ours)}  & \textbf{1.16}  & \textbf{0.63}   & 0.86           & $\mathbf{213}$   \\ \hline
\end{tabular}
\caption{Evaluation results of different trajectories and policies for extrinsic calibration with known intrinsic ground truth.}
\label{tab:extrinsic}
\end{table*}

\begin{table*}[!htp]
\footnotesize
\centering
\setlength{\belowcaptionskip}{-0.5cm}
\begin{tabular}{c|cccc}
\hline
\textbf{Joint calibration}     & \textbf{Mean error {[}\%{]}} & \textbf{Error std {[}\%{]}} & \textbf{Total path length {[}m{]}} & \textbf{Calibration time {[}s{]}}\\ \hline
random trajectory              & 9.48                         & 2.14                        & 2.26 &  $1975$                             \\
hand-crafted trajectory (long)        & 3.62                         & 1.31                        & 4.05  & $3167$                             \\
hand-crafted trajectory (short)       & 4.34                         & 9.59                        & 3.34  & $617$                             \\
model-based learned trajectory & 7.29                         & \textbf{1.28}               & 3.24 & $889$                              \\
random moving policy           & 5.82                         & 5.51                        & 1.16  & $613$                             \\
\textbf{model-free learned policy (ours)}      & \textbf{3.50}                & 2.26                        & \textbf{0.86}            &  $\mathbf{252}$          \\ \hline
\end{tabular}
\caption{Evaluation results of different trajectories and policies for intrinsic and extrinsic calibration.}
\label{tab:extrinsic_intrinsic}
\end{table*}

\subsection{Evaluation on Real Platform}
\label{sec:experiments-real}
With satisfactory results in the simulation, we continue to further evaluate our learned policy on the real platform.
We use a tightly time-synchronized \ac{vi} sensor system including an ADIS IMU and a global shutter fisheye camera~\cite{nikolic2014synchronized}.
We calibrate the intrinsic of the monochrome camera and extrinsic between the fisheye camera and the IMU sensor.
To evaluate our approach, we fix it on the end-effector of a real FRANKA EMIKA robot arm and run the learned policy online to generate a calibration trajectory, as is shown in Figure~\ref{fig:real}.

We add an additional automatic adjustment stage before each episode when the robot tries to align the center of the image view and the center of the target board.
Given a real \ac{vi} sensor, we directly run the learned policy to generate different motions according to the current state sequentially and record the generated trajectory.
%
%
The recorded data is used both for camera intrinsic calibration and \ac{vi} extrinsic calibration, in the same way as the simulation experiment.
The results are compared with the calibration results done by hand by a human expert. 
%
%
\\
\begin{table}[!hbt]
\footnotesize
\centering
\vspace{-0.4cm} 
\setlength{\belowcaptionskip}{-0.2cm}
\begin{tabular}{c|c|c}
\hline
{\textbf{Intrinsic}}  & {\textbf{Hand-crafted }} & {\textbf{Learned}} \\            
\hline
{fx}                                       & {468.713} & {468.253}\\
{fy}                                       & {468.748} & {468.327}\\
{cx}                                       & {364.419}& {364.912} \\
{cy}                                       &{215.564} & {215.813}\\
{k1}                                    & {0.005}& {0.011}\\
{k2}                                     & {-0.006}& {-0.053}\\
{k3} &{0.022} &  {0.153}\\
{p} & {-0.018} & {-0.135}\\ 
\hline
\end{tabular}
\caption{Comparison of intrinsic calibration results between hand-crafted trajectory and learned trajectory.}
\label{tab:real_int}
\end{table}

\begin{table}[!hbt]
\footnotesize
\centering
\vspace{-0.2cm} 
\setlength{\belowcaptionskip}{-0.3cm}
\begin{tabular}{c|cc}
\hline
\textbf{Extrinsic} & \textbf{Hand-crafted} & \textbf{Learned}\\
\hline
roll {[}rad{]} & 1.565 & 1.562\\
pitch {[}rad{]} & 1.567 & 1.563\\
yaw {[}rad{]} & -0.004  & -0.001\\
x {[}m{]} & -0.0374 & -0.0406\\
y {[}m{]} & 0.0015  & -0.0087\\
z {[}m{]} & 0.0003 & -0.0001\\ 
\hline

\end{tabular}
\caption{Comparison of extrinsic calibration results between hand-crafted trajectory and learned trajectory.}
\label{tab:real_ext}

\end{table}

A comparison of intrinsic and extrinsic calibration results is shown in Table~\ref{tab:real_int} and Table~\ref{tab:real_ext} respectively.
For both intrinsic and extrinsic calibration, our calibrated results are similar to the handcrafted results obtained by an expert.
The comparison of calibration metrics is shown in Table~\ref{tab:real_output}, further proving that our proposed method and the human expert produce similar quality results.
Also in this experiment our policy is able to perform the calibration faster than the expert, requiring only a third of the time.
These experiments prove the transferability of our method for unified intrinsic and extrinsic calibration in the real case.
%
%
%
%

\begin{table}[!t]
\footnotesize
\centering
\vspace{0.7em} 
\setlength{\belowcaptionskip}{-0.6cm}
\scalebox{1.0}{
\begin{tabular}{c|c|cc}
\hline
\textbf{Tasks} & \textbf{Outputs}  & \textbf{Hand-crafted}& \textbf{Learned}\\
\hline
\multirow{2}{*}{\textbf{intrinsic}} & calibration time {[}s{]}& 562 & 186\\
&reproj err {[}px{]} & 0.045&0.030 \\
\hline
\multirow{4}{*}{\textbf{extrinsic}} & calibration time {[}s{]}& $2221$& 747\\    
& reprojection error {[}px{]}&0.053 & 0.045\\
& gyroscope error {[}rad/s{]}& 0.004& 0.004\\
& accelerometer error {[}$\mathrm{m/s^2}${]}& 0.028& 0.094\\
\hline
\end{tabular}}
\caption{Comparison of calibration outputs between hand-crafted trajectory and learned trajectory.}
\label{tab:real_output}
\end{table}

\section{CONCLUSION}
\label{sec:conclusion}
In this work, we proposed the use of model-free deep \ac{rl} to learn a policy that guides the movement of a robotic arm holding a \ac{vi} sensor to efficiently collect measurements that can be used for both camera intrinsic calibration and \ac{vi} extrinsic calibration.
For joint intrinsic and extrinsic calibration our proposed method outperforms all other benchmarks in terms of both lowest mean calibration error and shortest path length.
Evaluations on the simulation platform and real robot system show that our learned policy is able to generate better motion primitives than human experts, leading to a much faster calibration process.
In simulation our proposed method is more than $10\times$ faster than a hand-crafted trajectory, which in our real-world experiment transfers to a speed-up of $3\times$ over humans.
Our method is ideally suited for a factory scenario where high throughput calibration is needed, or as a learning tool to teach optimal calibration motions to human experts.

%

%
%
%
In future work, the use of an Aprilgrid~\cite{olson2011apriltag} could enhance observability as it allows for different viewing angles.
In addition, partially visible Aprilgrid targets can be detected without problems, which would greatly simplify the data collection process. 
%
%


\clearpage

\bibliographystyle{./IEEEtran} 
\bibliography{./IEEEabrv.bib,./reference.bib}

\end{document}